\documentclass[conference]{IEEEtran}
\IEEEoverridecommandlockouts
\usepackage{cite}
\usepackage{amsmath,amssymb,amsfonts}
\usepackage{algorithmic}
\usepackage{graphicx}
\usepackage{textcomp}
\usepackage{color}
\usepackage{comment}
\def\BibTeX{{\rm B\kern-.05em{\sc i\kern-.025em b}\kern-.08em
    T\kern-.1667em\lower.7ex\hbox{E}\kern-.125emX}}
    
\usepackage{epsfig}
\usepackage{siunitx}
\usepackage{subfig}
\usepackage{mathtools}
\usepackage{booktabs}       
\usepackage{multirow}
\usepackage{bbm}
\usepackage{caption}
\usepackage[table,xcdraw]{xcolor}
\usepackage{pifont}
\usepackage{mwe}
\usepackage{cuted}
\usepackage{lipsum}
\usepackage[switch]{lineno}

\newcommand{\fref}[1]{Fig.~\ref{#1}}
\newcommand{\sref}[1]{Section~\ref{#1}}
\newcommand{\tref}[1]{Table~\ref{#1}}

\newcommand{\xmark}{\ding{55}}
\newcommand{\cmark}{\ding{51}}
\begin{document}

\title{Pseudo Supervised Monocular Depth Estimation with Teacher-Student Network
\author{Huan Liu\textsuperscript{1}, Junsong Yuan\textsuperscript{2}, Chen Wang\textsuperscript{3}, and Jun Chen\textsuperscript{1}
\thanks{\textsuperscript{1}Huan Liu and Jun Chen are with the Department of Electrical and Computer Engineering, McMaster University, ON L8S 4L8, Canada. {\tt\small \{liuh127, junchen\}@mcmaster.ca}}
\thanks{\textsuperscript{2}Junsong Yuan is with the Computer Science and Engineering Department, University at Buffalo, NY 14260, USA.  {\tt\small jsyuan@buffalo.edu}}
\thanks{\textsuperscript{3}Chen Wang is with the Robotics Institute, Carnegie Mellon University, Pittsburgh, PA 15213, USA. {\tt\small chenwang@dr.com}}
}}

\maketitle

\begin{abstract}
Despite recent improvement of supervised monocular depth estimation, the lack of high quality pixel-wise ground truth annotations has become a major hurdle for further progress. In this work, we propose a new unsupervised depth estimation method based on pseudo supervision mechanism by training a teacher-student network with knowledge distillation.  It strategically integrates the advantages of supervised and unsupervised monocular depth estimation, as well as unsupervised binocular depth estimation. Specifically, the teacher network takes advantage of the effectiveness of binocular depth estimation to produce accurate disparity maps, which are then used as the pseudo ground truth to train the student network for monocular depth estimation.
This effectively converts the problem of unsupervised learning to supervised learning. Our extensive experimental results demonstrate that the proposed method outperforms the state-of-the-art on the KITTI benchmark.
\end{abstract}


\section{Introduction}\label{introduction}




Estimating depth from a single image is a challenging but valuable task in both computer vision and robotics. 
Recently, we have witnessed the tremendous success of monocular depth estimation in assisting complicated computer vision tasks such as 3D scene reconstruction, visual optometry \cite{tateno2017cnn}, and augmented reality \cite{ramamonjisoa2019sharpnet}.
This success can be largely attributed to large-scale labeled datasets and deep convolutional neural network (DCNN) models. 
However, it can be very costly and in some cases impossible to obtain pixel-wise ground truth annotations for supervised training.
As such, great attention has been paid to unsupervised monocular depth estimation \cite{left-right,refine-distill,ego-motion,towards-scene} in recent years. A common approach is to formulate unsupervised monocular depth estimation as a self-supervised image reconstruction problem \cite{left-right,garg2016unsupervised}.

Despite its innovativeness, this approach has two intrinsic weaknesses. 
1) Compared to the supervised monocular setting, they often use the photometric loss to indirectly control the quality of disparity maps, which is less effective.
2) Compared to the unsupervised binocular setting, using one image to generate the disparity map (with the second image indirectly involved) is less effective than simultaneously exploiting the stereo pairs. Intuitively, the two weakness are intimately related to the nature of unsupervised and monocular approach and consequently inevitable. In this work, we aim to train an unsupervised monocular depth estimation network that can partially avoid these weaknesses by using a teacher-student based pseudo supervision for monocular depth estimation.

\begin{figure}[!t]
	\centering
	\includegraphics[width=1.0\linewidth]{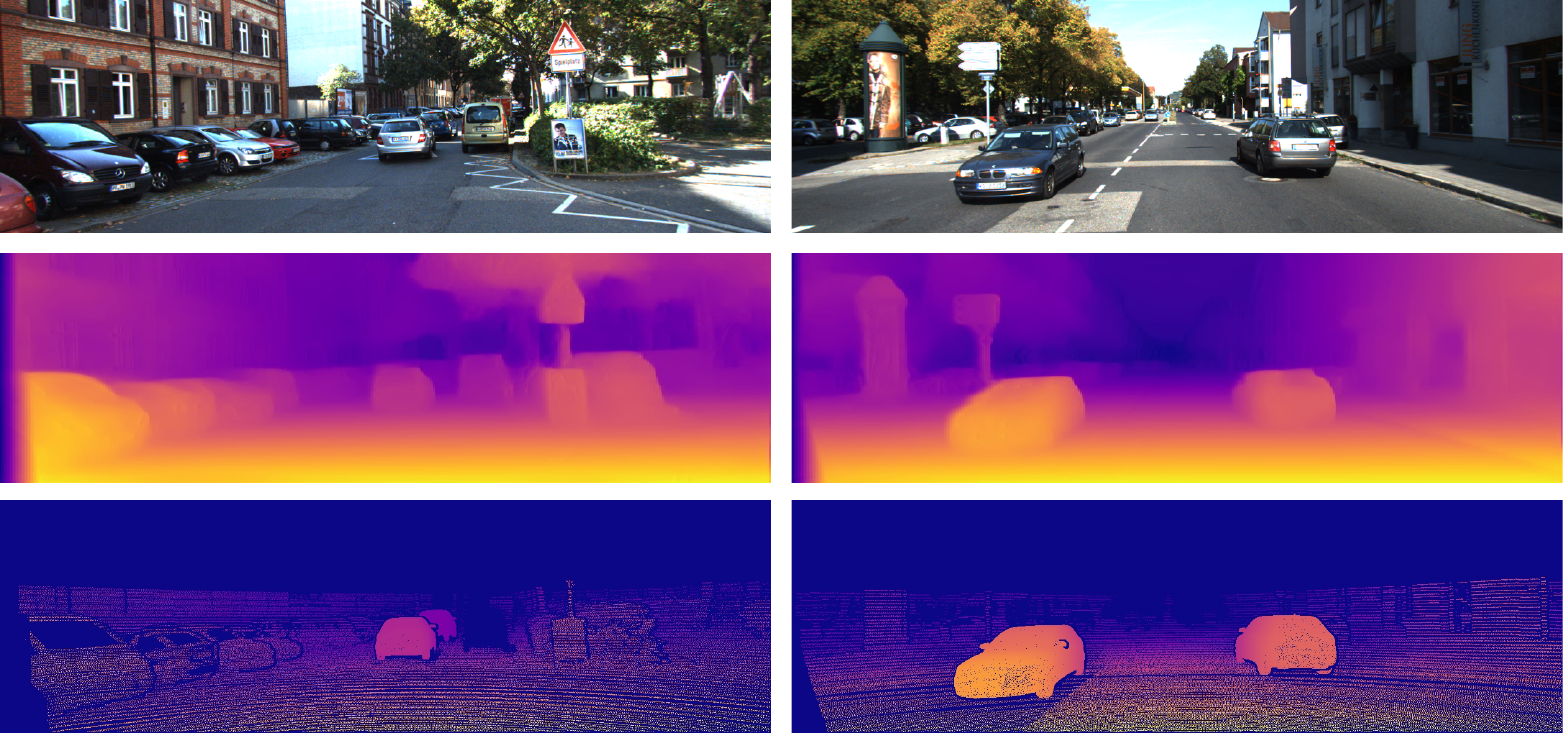}
	\caption{Example of the depth estimation results on KITTI 2015 stereo 200 training set \cite{Kitti} by our proposed pseudo supervision mechanism. From the top to bottom are respectively the input images, our results and sparse ground truth disparities. }
	\label{begin}
\end{figure}

To this end, we propose a novel pseudo supervision scheme, which
is leveraged to train the teacher-student network with distillation \cite{distillation}.
Specifically, the teacher network takes advantage of the effectiveness of unsupervised binocular depth estimation to produce accurate disparity maps.
The disparity maps are then used as the pseudo ground truth to train the student network for monocular depth estimation, which converts the problem of unsupervised learning to supervised learning. 
This pseudo supervision mechanism enables us to exploit the benefits of both supervised learning and binocular processing for unsupervised monocular depth estimation. As a consequence, the aforementioned two weakness can be tackled to a certain extent. 

However, 
in view of that it is not always possible to achieve perfect performance for the teacher network due to occlusion \cite{zhou2017unsupervised}, in the distillation process the student network is also provided with occlusion maps, which indicate the performance gap between the teacher network's prediction (pseudo ground truth for the student) and the real ground truth. 
This occlusion indication allows the student to focus on dealing with the un-occluded regions.
Moreover, the depth predictions in occlusion region still need to be carefully handled.
To address this problem, we train the teacher network with semantic supervision to enhance the performance around the occlusion boundaries, which was verified to be effective  \cite{ladicky2014pulling,eigen2015predicting,wang2015towards,towards-scene}.

The main contributions of this work can be summarized as follows.
1) By taking advantages of both unsupervised binocular depth estimation and pseudo supervised monocular depth estimation, we propose a novel mechanism for unsupervised monocular depth estimation. 
2) We fuse both occlusion maps and semantic representations wisely to handle the occlusion problem as well as boost the performance of student network.
3) We demonstrate through extensive experiments that our method outperforms the state-of-the-arts both qualitatively and quantitatively on the benchmark dataset\cite{Kitti}.


\section{Related Works}

The existing monocular depth estimation methods can be roughly divided into two categories.

\paragraph{Supervised / Semi-supervised Monocular Depth Estimation}
Supervised monocular depth estimation has been extensively studied in the past years. In the deep-learning framework, the problem becomes designing a neural network to learn the mapping from the RGB inputs to the depth maps. Eigen \textit{et al.} \cite{eigen2014depth} proposed a two-scale structure for global depth estimation and local depth refinement.  
Laina \textit{et al.} \cite{laina2016deeper} and Alhashim \textit{et al.} \cite{alhashim2018high} showed that better depth estimation results can be achieved with more powerful designs based on ResNet \cite{he2016deep} and DenseNet \cite{huang2017densely}.
There are also some works exploring the possibility of boosting the mapping ability of neural networks using statistical learning techniques. For example, Roy \textit{et al.} \cite{regression_forest} considered the combination of regression forests and neural networks; 
 \cite{crf1,crf2,crf3,crf4} used conditional random fields (CRFs) and CNNs to obtain sharper depth maps with clear boundary.

Due to their alleviated reliance on large labeled real-world datasets, semi-supervised methods have also received significant attention. Nevertheless, they still require some additional information \cite{semi-ordinal1,semi-ordinal2,semi-size1}.  In particular, Guo \textit {et al.} \cite{stereo-distill} proposed a teacher-student network for depth estimation, where the teacher network is trained in a supervised manner, albeit largely with synthetic depth data, and its knowledge is then transferred to the student network via distillation. Our work is partly motivated by the observation that the teacher network can actually be trained in a completely unsupervised manner without relying on any ground truth depth information (not even those associated with  synthetic images). 

\begin{figure}[!t]
	\centering
	\includegraphics[width=0.9\linewidth]{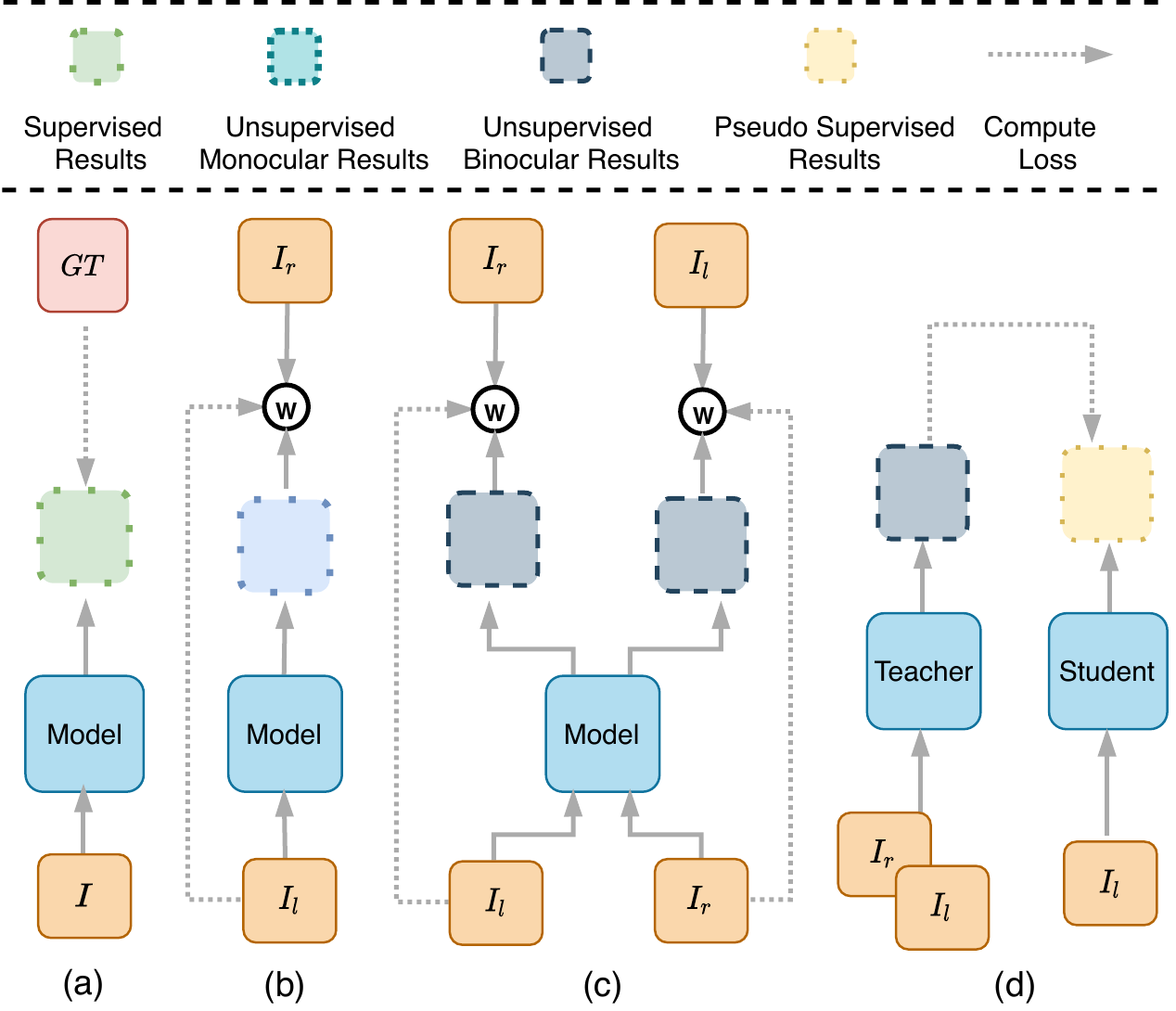}
	\caption{We show the architectures of (a) supervised/ (b) unsupervised monocular depth estimation, (c) unsupervised binocular depth estimation, and (d) our pseudo supervised mechanism.}
	\label{fig:formulation}
\end{figure}

\paragraph{Unsupervised Monocular Depth Estimation}
In the unsupervised setting, only the RGB domain information, typically in the form of stereo images or video sequences, is provided. Many training schemes and loss functions have been proposed for unsupervised depth estimation to exploit photometric warps. Garg \textit{et al.} \cite{garg2016unsupervised} constructed a novel differentiable inverse warping loss function. 
Zhou \textit{et al.} \cite{9013050} proposed a windowed bundle adjustment framework with considering constraints from consecutive frames with clip loss.
Godard \textit{et al.} \cite{left-right}
introduced the notion of left-right consistency, which is imposed on both images and disparity maps. Other consistency requirements, such as trinocular consistency \cite{trinocular} and bilateral consistency \cite{bilateral}, were also investigated.  In addition, there have been various attempts to take advantage of  generative adversarial networks (GANs) \cite{gan1,gan2,semi-geometry-tao}, knowledge distillation ~\cite{refine-distill}, synthetic datasets \cite{semi-geometry-tao,semi-syn2,semi-syn3,semi-syn4}, or semantic information  \cite{towards-scene,9196723, scene1,scene2,scene3}. 
Among them, arguably most relevant to the present paper is \cite{refine-distill}, where Pilzer \textit{et al.} proposed a distillation mechanism based on the concept of  cycle inconsistency. However, their adopted network structure is not very effective in simultaneously exploring the stereo pair  and suffers from a mismatching problem \cite{towards-scene}. In contrast, it will be seen that the proposed approach can take advantage of the efficiency of binocular processing in the training phase. Many recent works have recognized the benefit of exploiting semantic information for depth estimation via multi-task learning. Common approaches \cite{9196723, scene1,scene2,scene3} to multi-task learning typically involve neural networks with sophisticated structures. In contrast, Chen \textit{et al.} \cite{towards-scene} showed that it suffices to use a simple encoder-decoder network with  a task identity variable embedded in the middle. Inspired by \cite{conditional_gan}, we propose an alternative implementation with the task label stacked to the input images from the semantic dataset and KITTI to guide the teacher network for multi-task learning.

\section{Proposed Method} 
\subsection{Pseudo Supervised Depth Estimation Formulation} \label{analysis}
In this section, we provide a systematic comparison of several existing depth estimation formulations and show how the proposed pseudo supervision mechanism strategically integrates the desirable characteristics of different formulations.

\paragraph{Supervised Monocular Depth Estimation}
Let $I$ and $h_{gt}$ denote the input RGB image and its ground truth depth map, respectively.  Supervised training for monocular depth estimation aims to find a mapping $F$ that solve the following optimization problem (\fref{fig:formulation} (a)):   
\begin{equation}
\begin{aligned}
\arg\min_{F} \quad &  error(h_e, h_{gt}),\\
\mbox{s.t.}\quad
&h_e = F(I),
\end{aligned}\label{eq:supervised}
\end{equation}
where $h_e$ is the estimated depth map of $I$. Given a well-specified depth target, it is possible to train a DCNN model $\hat{F}_1$, as an approximate solution to (\ref{eq:supervised}), that is capable of lifting $I$ into a close neighborhood of $h_{gt}$. However, it can be very costly to obtain enough pixel-wise ground-truth annotations needed to specify the depth domain.

\paragraph{Unsupervised Depth Estimation}
The unsupervised depth estimation can be classified as monocular  and binocular depth estimation (stereo matching).
Due to the unavailability of a directly accessible depth map, the following formulations are often considered (\fref{fig:formulation} (b) and (c)):
\begin{equation}
\begin{aligned}
\arg\min_{F} \quad &  error(I_{el}, I_{l}), \\
\mbox{s.t.}\quad
&I_{el} = \langle I_r  \rangle_{d_l}, \ d_l = F(I_l), 
\end{aligned}
\label{unsupervised momo}
\end{equation}

\begin{equation}
\begin{aligned}
\arg\min_{F_l, F_r} \quad &  error(I_{el}, I_{l}) + error(I_{er}, I_{r}),\\
\mbox{s.t.}\quad
&I_{el} =  \langle I_r \rangle_{d_l}, \ d_l=F_l(I_l, I_r),\\
&I_{er} =  \langle I_l \rangle_{d_r}, \ d_r=F_r(I_l, I_r).
\end{aligned}\label{eq:binocular}
\end{equation}
where \eqref{unsupervised momo} and \eqref{eq:binocular} respectively refer to monocular and binocular estimation.
$(I_l, I_r)$ is a stereo pair, $\langle . \rangle$ is the warping operator,  and $d_{l(r)}$ denotes the estimated left (right) disparity map.
Note that $d_{l(r)}$ can be easily translated to a depth estimate given the focal length and the camera distance. 

However, these solutions are in general not as good as $\hat{F}_1$ for the following reasons : 1) Using the warped image $I_{el(er)}$ with respect to $I_{l(r)}$ to indirectly control the quality of the depth estimate is less effective than comparing the depth estimate directly with the ground truth as done in the supervised setting.
2) $I_l$ and $I_r$ often exhibit slightly different object occlusion, rendering perfect estimation of $d_{l(r)}$ impossible.
Nevertheless, $\hat{F}_3$ in principle performs better than $\hat{F}_2$   since monocular processing can be viewed as a degenerate form of binocular processing. 
Of course, the necessity of using stereo pairs as inputs restricts the applicability of binocular depth estimation.

\paragraph{Pseudo Supervision Mechanism} To strategically integrate the desirable characteristics of supervised monocular depth estimation, unsupervised monocular depth estimation, and unsupervised binocular depth estimation, we propose a pseudo supervision mechanism (\fref{fig:formulation} (d)) as follows:
\begin{equation}
\begin{aligned}
\arg\min_{F_s, F_t} \quad &  error(d_{e}, d_{\Tilde{gt}}),\\
\mbox{s.t.}\quad
&d_{e} = F_s(I_l), d_{\Tilde{gt}} = F_t(I_l, I_r),
\end{aligned}
\end{equation}
where $F_t$ is a teacher network and $F_s$ is a student network.
The teacher network trained  with stereo pairs $(I_l, I_r)$  as in \fref{fig:formulation} (c). Due to the advantage of binocular processing, the teacher network can be trained efficiently in an unsupervised manner and produce reasonably accurate disparity estimate.
The pseudo ground truth disparity maps $d_{\Tilde{gt}}$ produced by the trained teacher network $\hat{F}_t$ enable the student network to take advantage of supervised learning; moreover, in contrast to $\hat{F}_t$, the trained student network $\hat{F}_s$  is capable of performing monocular depth estimation. 
In order to ensure the pseudo ground truth produced by $\hat{F}_t$ with higher quality, 
a non-depth information (i.e. semantic maps) is integrated.
The detailed implementation of the pseudo supervision mechanism is described below.


\begin{figure*}[!t]
	\centering
	\includegraphics[width=0.9\linewidth]{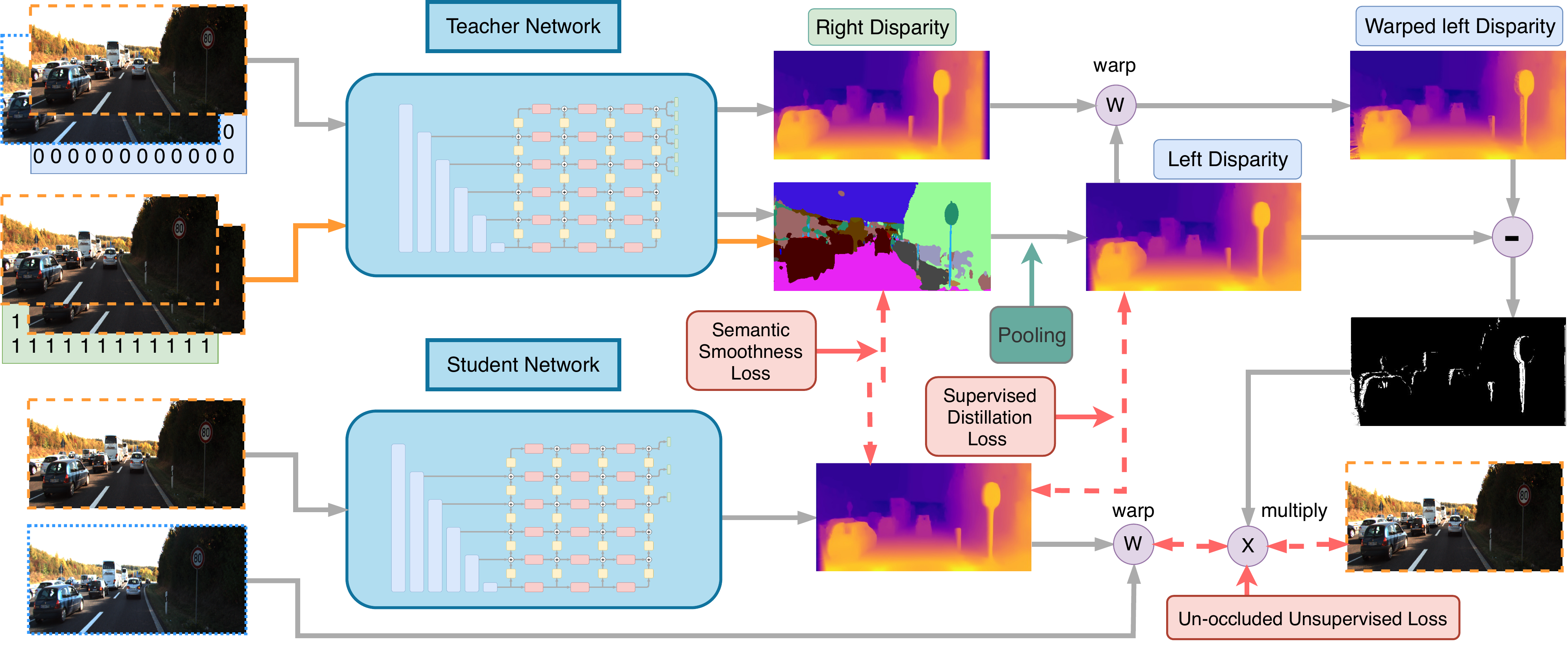}
	\caption{The pipeline of our proposed pseudo supervision mechanism. The teacher network is trained with alternating task-specific inputs ($0$ for semantic segmentation and $1$ for depth estimation) while the student network is trained using the pseudo ground truth. During inference, the student take a single image and produce its disparity map accordingly. }
	\label{network}
\end{figure*}

\subsection{Training the Teacher Network} \label{teacher_training}
The teacher network is designed to thoroughly exploit the training data and provide the pseudo ground truth to the student network (see \fref{network}). In addition, the teacher network is trained to learn the semantic information as well. 




\paragraph{Depth Estimation with Semantic Booster} \label{teacher_loss}
Most depth estimation methods exploit semantic information by employing a two-branch network where semantic segmentation and depth estimation are performed separately. In contrast, inspired by \cite{towards-scene} and \cite{conditional_gan}, we design an encoder-decoder network that can switch between the aforementioned two tasks according to a task label. Given the input images $I$ and the associated task labels $c$, the network outputs a task-specific prediction $Y = F_t(I, c)$. We set  $c = 0$ when the network is trained for depth estimation and set $c=1$ when the network is trained for semantic segmentation. 

For semantic segmentation, we train our network supervised with ground truth semantic maps from an urban scene dataset. The loss function $\mathcal{L}_{seg}$ for this task is:
\begin{equation}
\mathcal{L}_{seg} = \mathcal{CE}(F_t(I, c=0), gt),
\label{seg}
\end{equation}
where $\mathcal{CE}$ denotes cross-entropy loss and $gt$ specifies the semantic ground truth label.

In contrast, for binocular depth estimation (\textit{i.e.}, when $c=1$), we adopt unsupervised training. Following \cite{left-right}, we formulate the problem as minimizing the photometric reprojection error (see \fref{fig:formulation}(c) and (\ref{eq:binocular})).  Specifically, given two views $I_l$ and $I_r$, the network predicts their corresponding disparity maps $d_l$ and $d_r$, which are used to warp the opposite views; the resulting $\tilde{I}_l\triangleq\langle I_r \rangle_{d_l}$ and $\tilde{I}_r\triangleq\langle I_l \rangle_{d_r}$ serve as the reconstructions of $I_l$ and $I_r$, respectively. 
The loss function is a combination of $L1$ loss and single scale \textbf{SSIM} \cite{ssim} loss:
\begin{equation}
\mathcal{L}_{re}(I, \tilde{I}) = \theta \frac{\textbf{1} - \textbf{SSIM}(I - \tilde{I})}{2} + (1-\theta) \|I - \tilde{I} \|_1,
\label{re}
\end{equation}
where $\theta$ is set to $0.5$, and \textbf{SSIM} uses a $3 \times 3$ filter. We also adopt the left-right consistency loss $\mathcal{L}_{lr}$ and the disparity smoothness loss $\mathcal{L}_{sm}$ introduced in \cite{left-right}:
\begin{align}
&\mathcal{L}_{lr}(d, \tilde{d}) = \| d -  \tilde{d} \|_1,\\
&\mathcal{L}_{sm}(d, I) = |\partial_x d |e^{- \| \partial_x I \|} + |\partial_y d|e^{- \| \partial_y I \|}, \label{smooth}
\end{align}
where $\tilde{d}_l=\langle d_r \rangle_{d_l}$, $\tilde{d}_r=\langle d_l \rangle_{d_r}$, and $\partial$ is the gradient operator. 
Therefore, the total loss for unsupervised binocular depth estimation is $\mathcal{L}_{bi}$:
\begin{equation}
\begin{aligned}
\mathcal{L}_{bi} (d_l, d_r, I_l, I_r) &= \alpha_1 (\mathcal{L}_{re}(I_l, \tilde{I}_l) + \mathcal{L}_{re}(I_r, \tilde{I}_r))\\
&+ \alpha_2(\mathcal{L}_{lr}(d_l, \tilde{d}_l) + \mathcal{L}_{lr}(d_r, \tilde{d}_r))\\
&+ \alpha_3(\mathcal{L}_{sm}(d_l, I_l) + \mathcal{L}_{sm}(d_r, I_r) ).
\end{aligned}
\end{equation}

Following \cite{towards-scene}, after the training process for semantic segmentation converges, we use semantics-guided disparity smooth loss within each segmentation mask to boost disparity smoothness especially on object boundaries. During training, we only predict semantic segmentation on $I_l$ to reduce the computation load. Unlike \cite{towards-scene}, our semantic-guided smooth loss $\mathcal{L}_{semantic}$ is a simple variant of \eqref{smooth}:
\begin{equation} \label{semantic}
\mathcal{L}_{semantic}(d_l, s_l) = \mathcal{L}_{sm}(d_l, s_l),
\end{equation}
where $s$ denotes the predicted semantic map.

The overall loss function for the teacher network can be defined as follows:
\begin{equation}
\begin{aligned}
\mathcal{L}_{teacher}(d_l, d_r, I_l, I_r, s_l) &= \gamma_1 \mathcal{L}_{bi} (d_l, d_r, I_l, I_r) \\
&+ \gamma_2 \mathcal{L}_{semantic}(d_l, s_l).
\end{aligned}
\end{equation}


\subsection{Training the Student Network} \label{student_training}
Now we proceed to discuss the training strategy for the student network as shown in \fref{network}. 

\begin{figure*}[!t]
	\centering
	\includegraphics[width=0.9\linewidth]{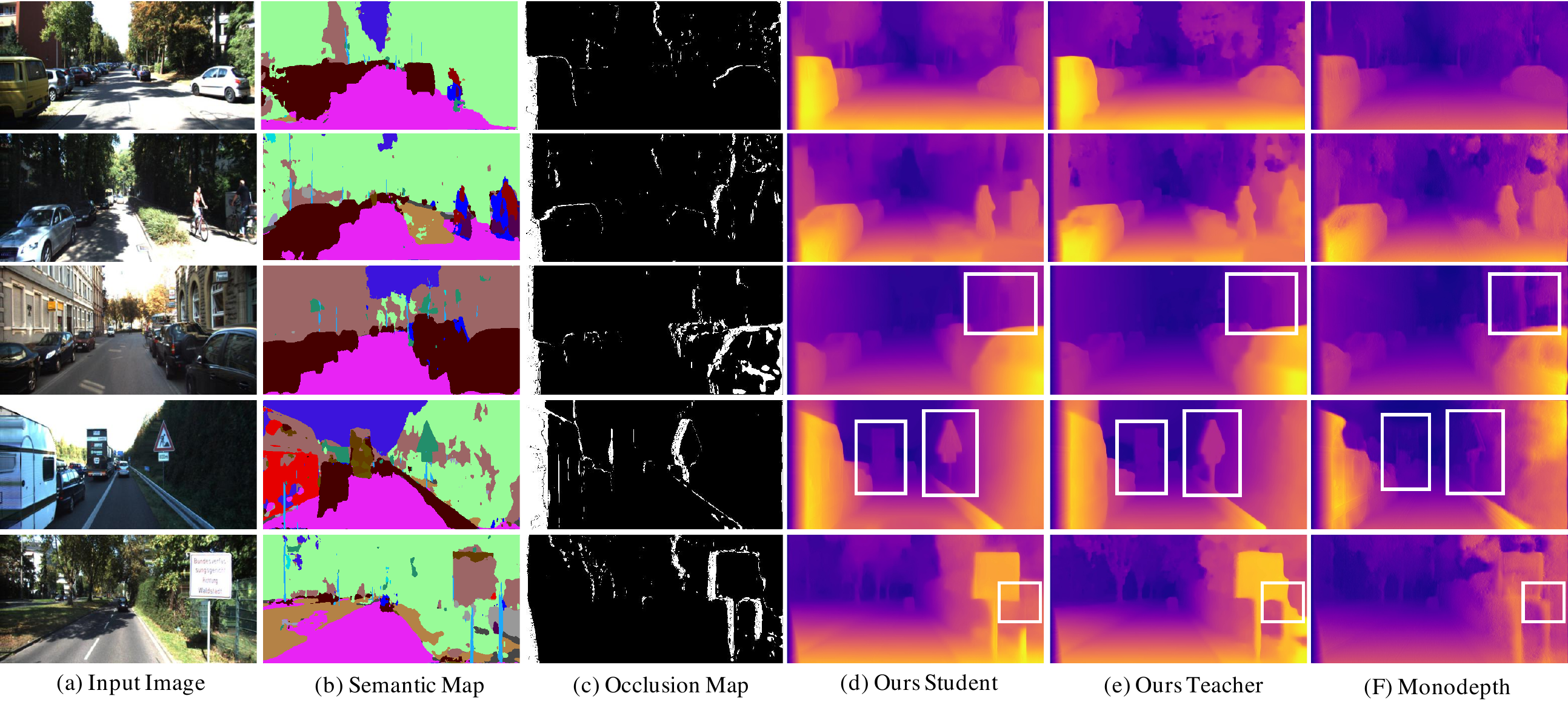}
	\caption{Illustrations of the experiment results on KITTI 2012 Eigen split \cite{eigen2014depth}. Monodepth denotes the results by Gordard \textit{et al.} \cite{left-right}. }
	\label{experiments}
\end{figure*}

\begin{table*}[!t]\small
	\centering
	\begin{tabular}{@{}c|c|c|c|c|c|c|c|c|c@{}}
		\toprule
		\multicolumn{1}{c|}{\multirow{2}{*}{Method}} & \multicolumn{1}{c|}{\multirow{2}{*}{Sup}} & 
		\multicolumn{1}{c|}{\multirow{2}{*}{Aux}} &
		 \multicolumn{4}{c|}{ \cellcolor[HTML]{FFCCC9} Error (lower, better)} & 
		\multicolumn{3}{c}{\cellcolor[HTML]{9AFF99} Accuracy (higher, better)} \\ 
		\multicolumn{1}{c|}{} & \multicolumn{1}{c|}{} & \multicolumn{1}{c|}{} & \multicolumn{1}{c|}{Abs Rel} & \multicolumn{1}{c|}{Sq Rel} & \multicolumn{1}{c|}{RMSE} & \multicolumn{1}{c|}{RMSE log} & \multicolumn{1}{c|}{$\delta < 1.25$} & \multicolumn{1}{c|}{$\delta < 1.25^2$} & \multicolumn{1}{c}{$\delta < 1.25^3$} \\ \midrule
		Eigen et al. \cite{eigen2014depth} & Y  & N & 0.203 & 1.548 & 6.307 & 0.282 & 0.702 & 0.890 & 0.958 \\
		Guo et al. \cite{stereo-distill} & Y  & D & \underline{0.096} & \underline{0.641} & \underline{4.059} & \underline{0.168} & \underline{0.892} & \underline{0.967} & \underline{0.986} \\
		Fu et al. \cite{fu2018deep} & Y  & N & \textbf{0.072} & \textbf{0.307} & \textbf{2.727} & \textbf{0.120} & \textbf{0.932} & \textbf{0.984} & \textbf{0.994} \\
		\midrule
		Garg \textit{et al.} \cite{garg2016unsupervised} & N  & N & 0.152 & 1.226 & 5.849 & 0.246 & 0.784 & 0.921 & 0.967 \\
		Pilzer et al. \cite{refine-distill}& N  & N & 0.142 & 1.231 & 5.785 & 0.239 & 0.795 & 0.924 & 0.968 \\
		Zhou et al. \cite{9013050} & N & N & 0.135 & 0.992 & 5.288 & 0.211 & 0.831 & 0.942 & 0.976 \\
		Gordard et al. (Monodepth) \cite{left-right} & N  & N & 0.124 & 1.388 & 6.125 & 0.217 & 0.841 & 0.936 & 0.975 \\
		
		Gordard et al. (Monodepth2) \cite{godard2019digging} & N  & N & \underline{0.115} & \underline{0.903} & \underline{4.863} & \underline{0.193} & \underline{0.877} & \underline{0.959} & \underline{0.981} \\
		\textbf{Ours (Student)} & N & N & \textbf{0.099} & \textbf{0.901} & \textbf{4.783} & \textbf{0.178} & \textbf{0.908} & \textbf{0.970} & \textbf{0.984} \\
		\midrule
		Chen et al. \cite{towards-scene} & N  & S & \underline{0.108} & \underline{0.875} & \underline{4.873} & \underline{0.204} & \underline{0.865} & \underline{0.956} & \underline{0.981} \\
		Lu et al.\cite{9196723}
		& N  & S & 0.115 & 1.202 & 5.828 & 0.203 & 0.850 & 0.944 & 0.980 \\
		
		\textbf{Ours (Student)}  & \textbf{N}  & S & \textbf{0.090} & \textbf{0.853} & \textbf{4.671} & \textbf{0.167} & \textbf{0.912} & \textbf{0.972} & \textbf{0.988} \\	
		\textbf{Ours (Teacher)}& \textbf{N}  & S & \textbf{0.059} & \textbf{0.777} & \textbf{3.868} & \textbf{0.137} & \textbf{0.959} & \textbf{0.983} & \textbf{0.991}\\
    \bottomrule
	\end{tabular}
	\caption{Quantitative comparison with state-of-the-art methods on the KITTI 2015\cite{Kitti} eigen split \cite{eigen2014depth}. Elements in the supervision (Sup) column are marked by yes (Y) or no (N) to describe whether the methods adopt a supervision manner. In the Auxiliary supervision (Aux) column, N represents 'no extra supervision', D stands for 'Depth supervision' and S denotes 'semantic supervision'. Best results are in \textbf{bold} and the second best are with 
	\underline{underline}. No matter if semantic information is used or not, our proposed method outperforms all the others.} 
	\label{quantitative}
\end{table*}

\paragraph{Supervised Training with Pseudo Disparity Ground Truth} \label{student_supervised}
The student network is trained under the supervision of the pseudo disparity ground truth provided by the teacher network. The adopted  pseudo supervised distillation loss $\mathcal{L}_{sup-mo}$ is an adaptation of the reconstruction loss \eqref{re} to disparity maps:
\begin{equation}
\begin{aligned}
\mathcal{L}_{sup-mo} (d_s, d_t)  = \mathcal{L}_{re}(d_s, d_t),
\end{aligned}
\end{equation}
where $d_s$  and $d_t$ are respectively the disparity estimate by the student and the  pseudo disparity ground truth from the teacher.

\paragraph{Unsupervised Training with Occlusion Maps}\label{student_occlusion}
Since the binocular teacher network naturally fails to find a good reconstruction in occlusion region\cite{zhou2017unsupervised}, the less capable monocular student network has little chance to succeed in this region. For this reason, it is sensible to direct the attention of the student network to other places where good reconstructions can be potentially found. Motivated by this, we generate an occlusion map from teacher as:
\begin{equation}
\mathcal{M}_{oc}(d, \tilde{d}) = \mathbbm{1}(|d - \tilde{d}| \leqslant 0.01),
\end{equation}
which sets the region that admits a good reconstruction (\textit{i.e.}, the region where the reconstructed $\tilde{d}$ is close to the original  map $d$) to 1 and sets the remaining part to 0. 

Based on occlusion map, we further define an un-occluded unsupervised loss $\mathcal{L}_{un-mo}$ by masking out the difficult region:
\begin{equation}
\begin{aligned}
\mathcal{L}_{un-mo} (d_s, I_s, \tilde{I}_s) &= \mathcal{M}_{oc} \mathcal{L}_{re}(I_s, \tilde{I}_s)
\end{aligned}
\end{equation}
where $\mathcal{L}_{re}$ and is the image reconstruction loss introduced in \sref{teacher_training} (a); $I_s$ and $\tilde{I}_s$ are respectively the monocular input and its reconstruction.

The semantic information $S_t$ from the teacher network is also used to guide the training of the student network  via loss \eqref{semantic} for handling occlusion boundaries. 
The total loss function for the student  network  can  be defined as follow:
\begin{equation}
\begin{aligned}
\mathcal{L}_{student} (I_s, \tilde{I}_s, d_s, d_t) &= \gamma_3 \mathcal{L}_{sup-mo} (d_s, d_t) \\
&+ \gamma_4 \mathcal{L}_{un-mo} (d_s, I_s, \tilde{I}_s) \\
& + \gamma_5\mathcal{L}_{semantic}(d_s, S_t).
\end{aligned}
\end{equation}

In the inference phase, the student network $F_s$ takes an image $I_s$ and produces a disparity $d_s = F_s(I_s)$, from which the depth estimate $D_s$ can be readily computed according to the formula
$D_s = bf/d_s$, where $b$ is the baseline distance between the cameras and $f$ is the focal length of lenses.

\section{Experiments}
\subsection{Implementation Details}
\paragraph{Network Architecture} \label{architecture}
As shown in \fref{network}, 
we shall refer to a specific encoder-decoder as Dense-Grid since the encoder is built using DenseNet161 \cite{huang2017densely} (in view of its feature extraction ability) without a linear layer while the decoder is built using  GridNet \cite{gridnet} (in view of its feature aggregation ability) with a shape of $6\times 4$.  
For the teacher network, the output end of each scale of the decoder is attached with  two $3\times 3$ convolutional layers. Depending on the task label, the first convolutional layer predicts semantic maps or left disparities (with the latter involving an extra global pooling step); the second convolutional layer predicts right disparities only. The two low-resolution disparity maps are up-sampled to full-scale to avoid texture-crop artifacts \cite{digging}. The structure of the student network is the same as that of the teacher network with the layers that predict segmentation and left disparities removed.

\paragraph{Regular Training Procedures and Parameters}
Our method is implemented using Pytorch \cite{paszke2017automatic} and evaluations are conducted on the Nvidia Titan XP GPU. Guided by alternating task labels, the teacher network is trained on KITTI \cite{Kitti} and Cityscape \cite{cordts2016cityscapes} for depth estimation and semantic segmentation. This training phase ends after 50 epochs when both tasks converge. The segmentation map produced in the last epoch of this training phase is leveraged to train the depth estimation task under total objective loss \eqref{semantic}.  With the pseudo ground truth and occlusion maps provided by the teacher network, the student network starts training process, which takes 50 epochs.

During training, inputs are resized to $256 \times 512$. Data augmentation is conducted as in Gordard \textit{et al.} \cite{left-right}.
We adopt the Adam optimizer with initial learning rate $\lambda = 10^4$, $\beta_1 = 0.9$, $\beta_2 = 0.999$, and $\epsilon = 10^5$. In the training of the student network the learning rate reduced at 30 and 40 epochs by a factor of 10, as well as the training of the teacher network. 
The weights of different loss components are set as following: $\gamma_1, \gamma_2, \gamma_3, \gamma_5, \alpha_1, \alpha_3  =1.0$,   $\gamma_4 =0.05$ and $\alpha_2=0.5$

\paragraph{Over-training of Teacher Network}\label{over}
Over-training is usually considered undesirable since it tends to jeopardize the generalization ability of a model. However, in our current context, it is actually desirable to train overly. Indeed, with over-training, the pseudo ground truth provided by the teacher network is likely to be very close to the actual ground truth of the training data (see \tref{overfitting}), which enables the student network to take advantage of pseudo supervised learning. 
Moreover, the fact that teacher network overfits the training data  has no impact on the generalization ability of the student network because we train our student regularly without over-training.
(Note that the generalization ability of the teacher is not a concern). To achieve this, we train our teacher network for depth task with additional 20 epochs. Without specifying, the student network performances reported in this paper are along with the over-trained teacher.

\begin{table}[!h]\small
\centering
\vskip -0.2cm
\captionsetup{font=small}
\scalebox{0.9}{
\begin{tabular}{c|c|c|c|c}
\toprule
Method & Abs Rel & Sq Rel & RMSE & RMSE log \\ \midrule
Teacher (over training) & \textbf{0.061} & \textbf{0.407} & \textbf{2.635} & \textbf{0.132} \\
Teacher (regular training)& 0.074 & 0.545 & 3.021 & 0.172 \\ \bottomrule
\end{tabular}}
\caption{Experimental results on KITTI 2012 Eigen split training set. Over-trained teacher can produce depth with lower error.}
\label{overfitting}
\end{table}

\subsection{Performance on KITTI}
Evaluations are conducted on KITTI 2012 and 2015 Eigen split \cite{eigen2014depth}. Evaluation metrics used in this work are the same as those in \cite{left-right} for fair comparison. 

\paragraph{Quantitative Results} \tref{quantitative} shows a quantitative comparison of several state-of-the-art depth estimation methods and the proposed one on KITTI 2015. 
Due to its binocular nature, the teacher network has a significant advantage over the monocular methods, which is clearly reflected in performance evaluations (the evaluation results of the teacher network reported in \tref{quantitative} are collected without over-training). Not surprisingly, the student network is less competitive than the teacher network; nevertheless, it still outperforms the other methods under comparison in terms of accuracy and error metrics. We additionally compare the performance of our proposed method with Guo \textit{et al.} \cite{stereo-distill}. For fair comparison, the model in \cite{stereo-distill} is trained with auxiliary ground truth depth and unsupervised fine-tuning on KITTI. Our student is trained with semantic maps (without ground truth depth). From \tref{guo_compare}, we can observe that without any supervision directly relevant to depth, our student still outperforms the Guo \textit{et al.} \cite{stereo-distill}. 

\begin{table}[!h]\small
\centering
\vskip -0.2cm
\captionsetup{font=small}
\scalebox{0.92}{
\begin{tabular}{c|c|c|c}
\toprule
Method & $\delta < 1.25$ & $\delta < 1.25^2$ & $\delta < 1.25^3$\\ \midrule
Guo et al. \cite{stereo-distill} (with depth) & 0.874 & 0.959 & 0.982 \\
Ours student (with semantic)& \textbf{0.912} & \textbf{0.972} & \textbf{0.988}  \\ \bottomrule
\end{tabular}}
\caption{Comparing with Guo \textit{et al.}. on KITTI 2015 eigen split. }
\label{guo_compare}
\end{table}

\paragraph{Qualitative Results} To further illustrate the effectiveness of the pseudo supervision mechanism, we show some qualitative results in \fref{experiments} on KITTI 2012. It can be seen that the disparity maps produced by the student network are comparatively the best in terms of visual quality and accuracy. For example, the edges of traffic signs and cars are clearer, and objects are detected with lower failure rate. It is also interesting to note that the disparity maps produced by the teacher network (which is over-trained) suffer from several problems (e.g.,  failure to  distinguish the traffic sign and the background in the last row of \fref{experiments}). That is to say, although the teacher network does not have a good generalization ability on the test dataset due to over-training, it is able to provide high-quality pseudo ground truth to train a student network.


\subsection{Ablation Study}

We perform ablation studies to demonstrate the effectiveness of each component in our proposed framework. Special attention is paid to three aspects: a) the benefit of incorporating semantic information in training the teacher, b) the advantage of joint utilization of pseudo ground truth (PGT), occlusion maps, and semantic information in training the student, c) inherent advantage of the proposed pseudo supervision mechanism.

\paragraph{Ablation Study for Training Teacher.}
We compare the cases with and without  semantic booster.  It can be seen from \tref{ablation} that the performance of the teacher network improves significantly with the inclusion of semantic information. 

\paragraph{Ablation Study for Training Student}
We consider using different combinations of pseudo ground truth (PGT), occlusion maps (Occ), and semantic information to train the student network. As shown by  \tref{ablation}, each element contributes positively to the performance of the student network, and the full combination outperforms any partial ones.

\paragraph{Inherent Advantage}
We re-implement our pseudo supervision mechanism using the ResNet-based structure proposed by Gordard \textit{et al.} \cite{left-right} in lieu of our Dense-Grid structure. It can be seen from \tref{ablation} that this re-implementation yields better performance as compared to the Monodepth network  \textit{et al.} with exactly the same  ResNet-based structure.
\begin{table}[!t]\small
	\centering
	
	\setlength{\tabcolsep}{0.4mm}{
	\begin{tabular}[c]{@{}c|c|c|c|c|c|c|c@{}}
		\toprule
		\multicolumn{1}{c|}{\multirow{2}{*}{Method}} & 
		\multicolumn{3}{c|}{Improvement} &  
		\multicolumn{4}{c}{\cellcolor[HTML]{FFCCC9}Error (lower, better)} \\
		\multicolumn{1}{c|}{} &
		\multicolumn{1}{c|}{PGT} & 
		\multicolumn{1}{c|}{Occ} & 
		\multicolumn{1}{c|}{Semantic} &
		\multicolumn{1}{c|}{Abs Rel} & 
		\multicolumn{1}{c|}{Sq Rel} & 
		\multicolumn{1}{c|}{RMSE} & 
		\multicolumn{1}{c}{RMSE log} 
		
		\\ \midrule
		
		\multirow{5}{*}{Student} 
		& \xmark &\xmark &\xmark & 0.127 & 1.215 & 5.520 & 0.268 \\
		& \cmark &\xmark &\xmark & 0.122 & 0.919 & 5.093 & 0.211 \\
		&  \cmark &  \cmark & \xmark& 0.119 & 0.959 & 5.056 & 0.210 \\
		&  \cmark &\xmark &  \cmark & 0.117 & 0.888 & \textbf{4.949} & 0.205\\
		&  \cmark &  \cmark &  \cmark & \textbf{0.115} & \textbf{0.885} & \textbf{4.956} & \textbf{0.202}  
		\\ \midrule
		\multirow{2}{*}{Teacher} 
		& \xmark& \xmark&  \xmark& 0.089  &0.973  &4.423  &0.190  \\
		&\xmark & \xmark&  \cmark  & \textbf{0.077} & \textbf{0.672} & \textbf{3.950} & \textbf{0.174} 
		\\ \midrule
		\multicolumn{4}{c|}{Monodepth Res50 Original} & 0.133  &1.142  &5.533  &0.230  \\
		\multicolumn{4}{c|}{Pseudo Supervised Monodepth} & 0.129  &1.112  & 5.236  & 0.217  \\
		\bottomrule
	\end{tabular}}
	
	\caption{Ablation studies on KITTI 2012 Eigen split \cite{eigen2014depth}. }
	
	\label{ablation}
\end{table}


\section{Conclusion}
In this paper, we propose a pseudo supervision mechanism to realize
unsupervised monocular depth estimation by strategically exploiting the benefits of supervised monocular depth estimation and unsupervised binocular depth estimation. We have also shown how to make effective use of  performance-gap indicator, and semantic booster in the implementation of the pseudo supervision mechanism. The experimental results indicate that the proposed unsupervised monocular depth estimation method performs competitively against the state-of-the-art. 
As to future work, apart from refining the proposed depth estimation method, we also aim to further enrich and strengthen the theoretical framework of pseudo supervision and explore its application to other computer vision problems. 

\clearpage

\bibliographystyle{unsrt}
\bibliography{egbib.bib}

\clearpage
\onecolumn

\begin{huge}
	\centering
	\textbf{Supplementary Material}
\end{huge}

\section{Network Architecture}

\begin{figure*}[!h]
	\centering
	\includegraphics[width=0.9\linewidth]{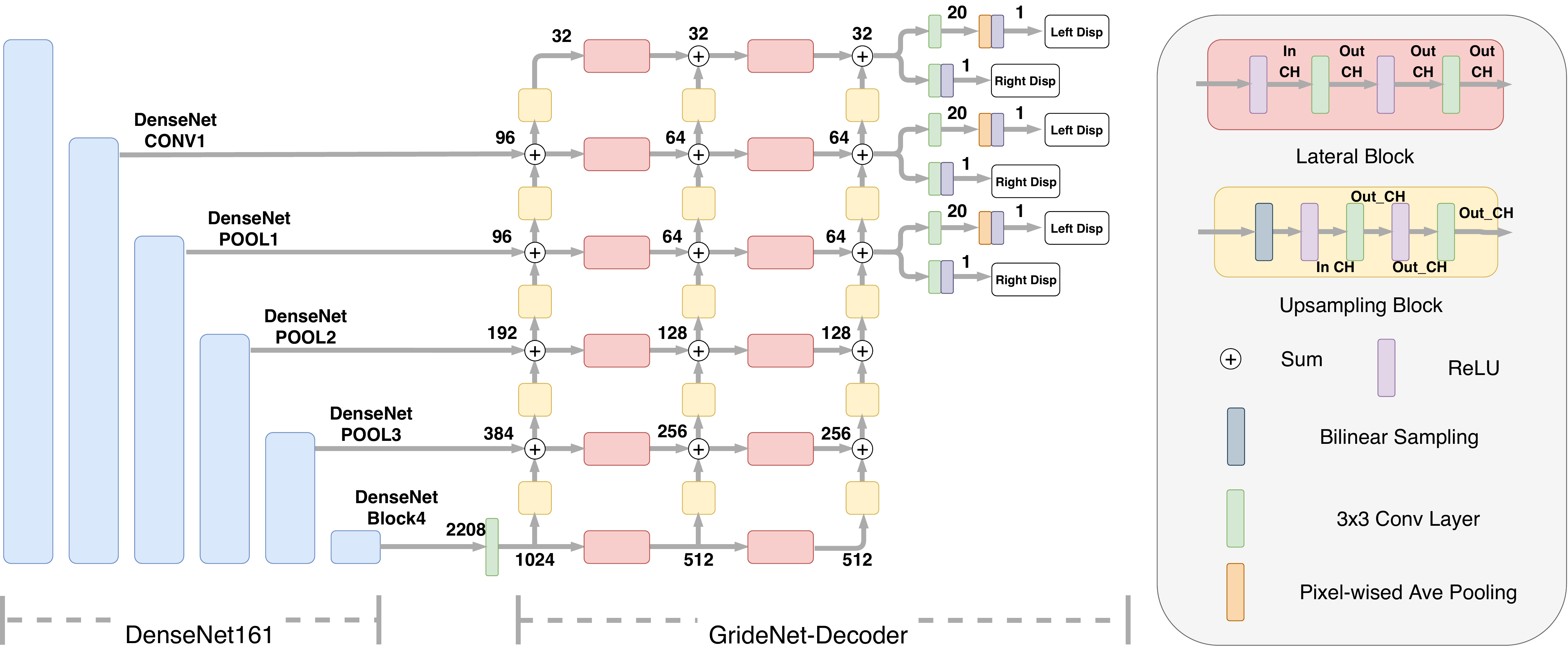}
	\caption{Architecture of our proposed Dense-GridNet. \textbf{In-CH} and \textbf{Out-CH} stand for input channels and output channels of the blocks or the convolutional layers. Specific numbers label the number of channels for each feature map. The output layers from DenseNet161 \cite{huang2017densely} is shown by their names defined in PyTorch \cite{paszke2017automatic} official model.}
	\label{archi_indetail}
\end{figure*}

Network architectures have been discussed in \sref{architecture} faithfully. We utilize the DenseNet161 \cite{huang2017densely} as encoder by removing the linear layer, while the decoder is built by GridNet \cite{gridnet} structure. Here, we show the architecture of our teacher in \fref{archi_indetail} for further explanations. Note that the architecture of the student is similar to the teacher except for removing the layers that predict segmentation and left disparities in the teacher. Batch normalization is not utilized in our network.

\section{Evaluation Metrics}
Here, we show our adopted evaluation metrics in detail. $d_i$ and $\hat{d}_i$ are respectively the ground truth disparity map and our estimate. $N$ is the total number of pixels in each image. \\

Mean relative error (Abs Rel): $\frac{1}{N} \sum_{i=1}^{N} \frac{\lVert \hat{d}_i - d_i \rVert}{d_i} $;
\ \ \ \ \ \ \ \ \ \ \ \ \ \ \ \ \ \ \ \ \ Square relative error (Sq Rel): $\frac{1}{N} \sum_{i=1}^{N} \frac{\lVert \hat{d}_i - d_i \rVert^2}{d_i} $; \\

Root mean square error (RMSE): $\sqrt{\frac{1}{N} \sum_{i=1}^{N} (\hat{d}_i - d_i )^2}$;\\

Mean $\log10$ square error (RMSE log): $\sqrt{\frac{1}{N} \sum_{i=1}^{N} \lVert\log\hat{d}_i - \log d_i \rVert^2}$; \\

Accuracy with threshold, $\delta < 1.25$, $\delta < 1.25^2$, $\delta < 1.25^3$, represent the percentage of $\hat{d}_i$ such that $\delta = max(\frac{d_i}{\hat{d}_i}, \frac{\hat{d}_i}{d_i}) < 1.25, 1.25^2$ or $1.25^3$ \\
\section{Aditional Evaluation Results}

\subsection{Qualitative Evaluation on Real-world Video}
Here we evaluate our proposed method on a real-world video shot in Singapore \footnote{https://www.youtube.com/watch?v=7LlXG8f5Hzo\&t=160s}. We select four clips from the video to achieve data diversity. The first and second clips record the urban view, third clip is captured in community, and fourth clip is taken on highway. Noted that the training data of KITTI is captured in Germany, which indicate there might be a domain gap between our training data and test video sequences. 
We also show the comparison with Gordard \textit{et al.} \cite{left-right} in our video. 
It can be observed that our method is more robust in real-world and generalize better than \cite{left-right}.


\subsection{Quantitative Results on KITTI 2015 }
Our evaluation are  conducted on the KITTI 2015 training set, which contains 200 high quality disparity maps with RBG images. Our model is trained on KITTI split. There are total 30,159 images in KITTI split, where we keep 29,000 for training and rest for validation.
The evaluation of both teacher and student are shown in \tref{quanti_supp}. As mentioned, the teachers here are trained to converge rather than over-fit on the dataset.

\begin{table*}[!h]\scriptsize
	\centering
	
	\begin{tabular}{@{}c|c|c|c|c|c|c|c|c@{}}
		\toprule
		\multicolumn{1}{c|}{\multirow{2}{*}{Method}} & \multicolumn{1}{c|}{\multirow{2}{*}{Training}} & 
		\multicolumn{4}{c|}{ \cellcolor[HTML]{FFCCC9} Error Metrics(lower, better)} & 
		\multicolumn{3}{c}{\cellcolor[HTML]{9AFF99} Accuracy Metrics(higher, better)} \\ 
		\multicolumn{1}{c|}{} & \multicolumn{1}{c|}{} &  \multicolumn{1}{c|}{Abs Rel} & \multicolumn{1}{c|}{Sq Rel} & \multicolumn{1}{c|}{RMSE} & \multicolumn{1}{c|}{RMSE log} & \multicolumn{1}{c|}{$\delta < 1.25$} & \multicolumn{1}{c|}{$\delta < 1.25^2$} & \multicolumn{1}{c}{$\delta < 1.25^3$} \\ \midrule

		Ours (Student)  & KITTI split & 0.106& 0.975 & 5.40 & 0.192 & 0.860 & 0.949 & 0.982\\
		Ours (Teacher)& KITTI split & 0.077& 0.672 & 3.950 & 0.174 & 0.924& 0.962 & 0.983\\
		
		\bottomrule
	\end{tabular}
	\caption{Results on KITTI 2015 \cite{Kitti} dataset. Elements in the Training column are marked by KITTI. Experiments are conducted capped at 80 meters in depth.}
	
	\label{quanti_supp}
\end{table*}

\subsection{Additional Qualitative Results}
\vspace{-0.3cm}
\begin{figure*}[htb]
	\centering
	\includegraphics[width=1.0\linewidth]{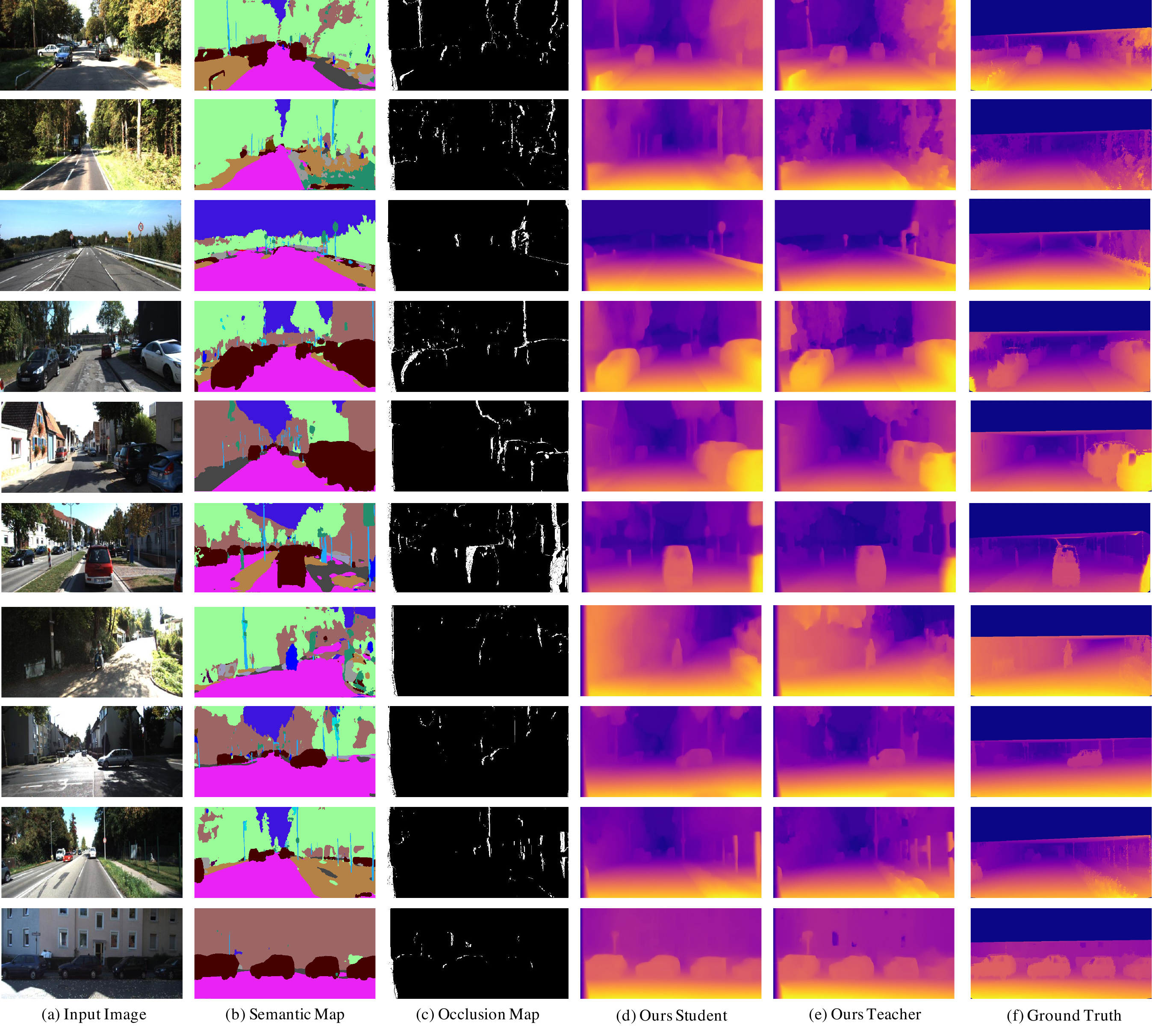}
	\caption{Illustrations of the experiment results on KITTI Eigen split test set \cite{Kitti} with a model trained on KITTI Eigen split \cite{eigen2014depth}, where the teacher network produces semantic maps and occlusion maps. We interpolate the extremely sparse ground truth for better visualization. }
	\label{experiments}
\end{figure*}

\begin{figure*}[!h]
	\centering
	\includegraphics[width=1.0\linewidth]{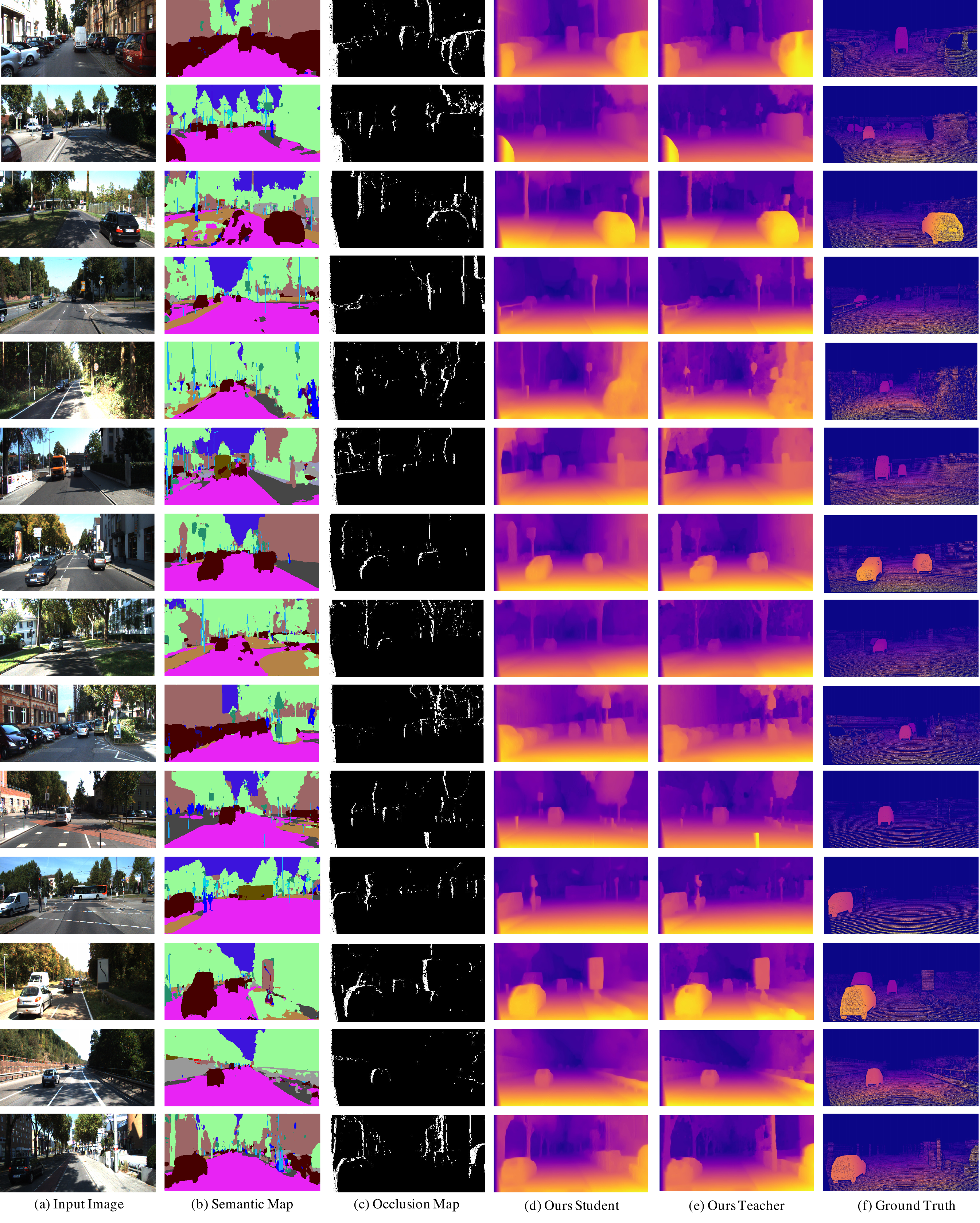}
	\caption{Illustrations of the experiment results on KITTI 2015  \cite{Kitti}  with a model trained on KITTI Eigen split \cite{eigen2014depth}, where the teacher network produces semantic maps and occlusion maps.}
	\label{experiments}
\end{figure*}

\end{document}